\documentclass[sigconf,screen]{acmart}

\AtBeginDocument{%
  }

\acmConference[ACM-BCB '24]{15th ACM International Conference on Bioinformatics, Computational Biology and Health Informatics}{November 22, 2024}{Shenzhen, Guangdong Province, P.R. China}

\RequirePackage{graphicx}
 \usepackage{booktabs}
\usepackage[most]{tcolorbox}
\usepackage{xcolor}
\usepackage{bm}
\usepackage{graphicx}
\usepackage{enumitem}

\usepackage{longtable}
\makeatletter
\def\set@curr@file#1{\def\@curr@file{#1}} 
\makeatother
\usepackage[load-configurations=version-1]{siunitx} 

\usepackage[mathlines, switch]{lineno}

\usepackage{booktabs}
\usepackage{rotating}

\usepackage{graphicx} 
\usepackage{booktabs} 
\usepackage{multirow} 

\title[TrialEnroll: Clinical Trial Enrollment Success Prediction]{TrialEnroll: Predicting Clinical Trial Enrollment Success with Deep \& Cross Network and Large Language Models}

\author{Ling Yue*}
\authornote{Both authors contributed equally to this research.}
\email{yuel2@rpi.edu}
\affiliation{%
  \institution{Rensselaer Polytechnic Institute}
  \city{Troy}
  \state{NY}
  \country{USA}
}

\author{Sixue Xing*}
\email{xings@rpi.edu}
\affiliation{%
  \institution{Rensselaer Polytechnic Institute}
  \city{Troy}
  \state{NY}
  \country{USA}
}

\author{Jintai Chen}
\email{cjt147@illinois.edu}
\affiliation{%
  \institution{University of Illinois Urbana-Champaign}
  \city{Chempaign}
   \state{IL}
  \country{USA}
}

\author{Tianfan Fu}
\email{fut2@rpi.edu}
\affiliation{%
 \institution{Rensselaer Polytechnic Institute}
 \city{Troy}
 \state{NY}
 \country{USA}
 }

\begin{document}

\begin{abstract}
Clinical trials need to recruit a sufficient number of volunteer patients to demonstrate the statistical power of the treatment (e.g., a new drug) in curing a certain disease. 
Clinical trial recruitment has a significant impact on trial success. 
Forecasting whether the recruitment process would be successful before we run the trial would save many resources and time. 
This paper develops a novel deep \& cross network with large language model (LLM)-augmented text feature that learns semantic information from trial eligibility criteria and predicts enrollment success. 
The proposed method enables interpretability by understanding which sentence/word in eligibility criteria contributes heavily to prediction. 
We also demonstrate the empirical superiority of the proposed method (0.7002 PR-AUC) over a bunch of well-established machine learning methods. 
The code and curated dataset are publicly available at \url{https://anonymous.4open.science/r/TrialEnroll-7E12}. 
\end{abstract}



\maketitle

\section{Introduction}


Clinical trials play an indispensable role in developing new treatments by assessing their safety and efficacy on human subjects~\cite{friedman2015fundamentals}. These trials serve as critical checkpoints in the drug development process, ensuring that medications are not only effective but also safe for public use~\cite{hill1952clinical}. To conduct a robust evaluation, it is imperative to enroll a sufficient number of participants who meet specific eligibility criteria~\cite{haddad2015barriers}. This ensures that the statistical power of the trial is adequate to detect any significant differences between the treatment group and the control group. 

However, recruiting the right number of participants is challenging~\cite{patel2003challenges,haddad2015barriers}. The process is often time-consuming and costly, which can delay the entire drug development timeline. One of the main reasons for this difficulty is the stringent eligibility criteria that must be met by potential participants~\cite{peto1978clinical}. These criteria are designed to ensure that the study population is representative of the intended patient population and to minimize confounding variables that could skew the results. 

To address this issue, there has been growing interest in leveraging machine learning algorithms to predict patient enrollment in clinical trials more accurately~\cite{lo2019machine,fu2023automated}. By analyzing historical data from previous trials, these algorithms can identify patterns and factors that influence recruitment rates. This predictive modeling can help researchers better plan and design their trials, leading to more efficient and effective recruitment strategies. 

However, the prediction of clinical trial enrollment encounters some data and technical challenges, as shown below. 
\begin{itemize}[leftmargin=*]
\item Lack of high-quality data. Clinical trial data are usually highly noisy and sensitive and not AI-ready, which hinders AI's deployment. 
\item Lack of ability to learn from the multimodal heterogeneous features. Clinical trials usually involve multimodal heterogeneous features, such as biomedical entities (drug, disease), and demographic features (e.g., gender and age). It is challenging for the current machine learning model to learn from them. 
\item Lack of ability to learn from unstructured text data. Clinical trial involves a great amount of unstructured text data. For example, eligibility criteria consist of multiple natural language inclusion and exclusion criteria, which specify what is wanted and unwanted during the patient recruitment process. It is challenging to capture semantic information from the unstructured text data. 
\end{itemize}

To address these challenges, we formally define the clinical trial enrollment prediction problem, curate AI-ready datasets, and customize Deep \& Cross Network~\cite{wang2017deep} using large language model-augmented features to learn semantic information from unstructured text data (e.g., eligibility criteria, LLM-generated feature), where large language model is used to augment the text feature of biomedical entities like drug and disease.

For ease of exposition, the major contribution of this manuscript can be summarized as follows. 
\begin{itemize}[leftmargin=*]
\item \textbf{Problem.} To the best of our knowledge, We are the first to identify clinical trial enrollment as a predictable problem and formulate it into an AI-solvable task. 
\item \textbf{Data.} We curate a ready-to-use dataset specialized for clinical trial enrollment prediction. The dataset contains 31,094 trials with binary labels for enrollment success.  
\item \textbf{Method.} We develop a deep \& cross network with large language model enhanced text feature tailored to enrollment prediction. Specifically, we design a hierarchical attention mechanism to learn the word- and sentence-level importance in an end-to-end manner. 
\item \textbf{Results.} We conduct extensive experiments to validate the effectiveness of the proposed method. Specifically, the proposed method obtains 0.7002 PR-AUC score and achieves 0.0229 improvement over the best baseline method. Also, our method exhibits desirable interpretability that could help clinicians understand the AI predictions. 
\end{itemize}

The rest of the paper is organized as follows. First, Section~\ref{sec:related} briefly reviews the related works in using AI for predictive clinical trial tasks. Then, we elaborate on our method in Section~\ref{sec:method}. After that, empirical studies are described in Section~\ref{sec:experiment}. Finally, we conclude the paper in Section~\ref{sec:conclusion}.

\section{Related Works}
\label{sec:related}

In this section, we discuss the works that use AI methods for automatic clinical trial planning and management from two perspectives: clinical trial problems and specific AI methodologies for these problems. 

\paragraph{AI-solvable clinical trial problems.}
AI, especially deep learning methods, has great potential in aiding many clinical trial problems. 
Specifically, they focus on the following clinical trial problems. 
\begin{itemize}[leftmargin=*]
\item Clinical trial outcome/approval prediction: The costs of conducting clinical trials are extremely expensive (up to hundreds of millions of dollars~\cite{martin2017much}), and the time to run a trial is very long (7-11 years on average) with low success probability~\cite{peto1978clinical,ledford20114}. However, many factors, such as the inefficacy of the drug, drug safety issues, and poor trial eligibility criteria design, can cause the failure of a clinical trial~\cite{friedman2015fundamentals}. If we were better at predicting the results of clinical trials, we could avoid running trials that will inevitably fail — more resources could be devoted to trials that are more likely to succeed. \cite{fu2022hint,lu2024uncertainty,chen2024uncertainty} propose to predict clinical trial approval based on drug molecule structure, disease code, and trial eligibility criteria. 
\item Patient-trial matching: Patient recruitment is a key step in running clinical trials. 
Given the trial's eligibility criteria, matching the appropriate patients based on their electronic health records is time- and labor-intensive~\cite{zhang2021ddn2,fu2024ddn3}. 
Patient-trial matching is formulated as a machine learning task to automate the process that selects appropriate patients for the target trial and alleviates the burden of patient recruitment. 
\cite{zhang2020deepenroll,gao2020compose} predict patient-trial matching based on trial eligibility criteria and patient electronic health records (EHR); 
\item Digital twin of clinical trial: Digital twins in the context of clinical trials refer to virtual representations of real-world patients or systems that can be used to simulate and predict outcomes under various conditions. Digital twins can simulate how different patient populations might respond to new treatments, potentially reducing the need for lengthy and costly physical trials. This can significantly speed up the drug development process. \cite{wang2024twin} design a TWIN-GPT model by finetuning standard GPT model to synthesize patient visit history to mimic the procedure of clinical trials and predict trial outcomes. 
\item Integration of multi-omics data. Multi-omics data enables the characterization of individual patients at a molecular level, which is crucial for precision medicine approaches. By understanding the genetic~\cite{lu2021cot,lu2022cot}, transcriptomic~\cite{lu2023deep}, and other molecular profiles of patients~\cite{chen2021data}, treatments can be tailored to match individual disease mechanisms, potentially leading to more effective and personalized therapies. 
\item Clinical trial duration prediction: Predicting the duration of clinical trials accurately offers significant benefits for trial management. By predicting trial duration, resource allocation such as staffing, budget, and facilities can be optimized, ensuring resources are available when needed to prevent inefficiencies and bottlenecks \citep{Kerali2018}.
\cite{yue2024trialdura} predicts clinical trial duration using textual information of various trial features (disease, drug, eligibility criteria) with a pretrained BioBERT~\cite{lee2020biobert} model as a text feature enhancement.
\item Clinical trial site selection: The site of the clinical trial, also known as the investigators, is the physical place where clinical trials are carried out and is the key to the success of clinical trials. 
The selection of clinical trial sites is complex and laborious work. Traditional ways usually depend heavily on human experts, who manually assign the clinical trial sites to the specific clinical trials. The process is time-consuming, error-prone, and expensive. To reduce the time, resources, and cost, \cite{srinivasa2022clinical} designed a policy-based reinforcement learning method to select trial sites automatically. 
\item Dataset: TrialBench~\cite{chen2024trialbench} identifies 8 AI solvable clinical trial problems (prediction of trial duration, patient dropout rate, serious adverse event, mortality rate, trial approval outcome, trial failure reason, drug dose-finding, design of eligibility criteria) and curates 24 AI-ready corresponding datasets to facilitate the AI for the clinical trial community. 
\end{itemize}

\paragraph{AI methodologies tailored to clinical trial.} 
Clinical trials produce valuable multimodal data that can be used for machine learning modeling. Patient-level trial data contain individual patient measurements and adverse events during the trial period. The trial summary data contains multi-modal information related to the trial, including drug molecules, target diseases, eligibility criteria for recruiting patients, sponsors (e.g., some specific pharmaceutical company or academic institute), physical trial sites (geographical locations to conduct the trial), trial aims and trial results. A series of deep learning methods were developed to represent these multimodal clinical trial features. For example, DeepEnroll~\cite{zhang2020deepenroll} also leveraged a Bidirectional Encoder Representations from Transformers (BERT~\cite{devlin2018bert}) model to encode clinical trial information. 
COMPOSE~\cite{gao2020compose} used pretrained BERT to generate contextualized word embedding for each word of the trial protocol and then applied multiple one-dimensional convolutional layers with different kernel sizes to generate trial embedding to capture semantics at different granularity levels. 
\cite{qi2019predicting} designs a Residual Semi-Recurrent Neural Network (RS-RNN) to predict the phase 3 trial results based on phase 2 results. 
There are a great deal of missing features in clinical trials. To handle this issue, 
\cite{lo2019machine} explored various imputation techniques~\cite{wu2022cosbin,lu2019integrated,lu2018multi} (such as mean imputation, random imputation, and k-nearest neighbor) and applied a series of conventional machine learning models (including logistic regression~\cite{lavalley2008logistic}, random forest~\cite{breiman2001random}, SVM~\cite{jakkula2006tutorial}) to predict the outcome of clinical trial within 15 disease groups. However, they do not consider drug molecule features and trial protocol information and thus could not differentiate the outcome for different drugs focusing on disease. 
\cite{fu2022hint,fu2023automated} designed a hierarchical interaction network (HINT) to encode multimodal trial features and capture their interaction (including drug molecules, disease code, and eligibility criteria). 
Based on this work, \cite{chen2024uncertainty,lu2024uncertainty} extend its scope by quantifying uncertainty and studying the explainability/interpretability of the HINT model. 
\cite{yue2024trialdura} design a hierarchical attention mechanism for learning word- and sentence-level semantic information from trial eligibility criteria. 
\cite{yue2024ct} design a multi-agent large language model-based reasoning method for clinical trial outcome prediction. Specifically, they decompose clinical trial outcome prediction into several simple sub-tasks, e.g., trial enrollment success, drug safety, and drug efficacy. 
To predict the clinical trial outcome, 
\cite{gao2024language} designed a large language model-based interaction network (LINT), which uses a large language model (LLM) to extract meaningful text embedding and provide fruitful features, followed by a small-scale interaction network (to be finetuned) to make the prediction. 
\cite{srinivasa2022clinical} design a policy-based reinforcement learning and design fairness-aware reward function to enhance the fairness of clinical trials over different races, especially for minority groups.

\section{Methodology}
\label{sec:method}

\begin{figure}[ht]
\centering
\includegraphics[width=0.5\textwidth,keepaspectratio]{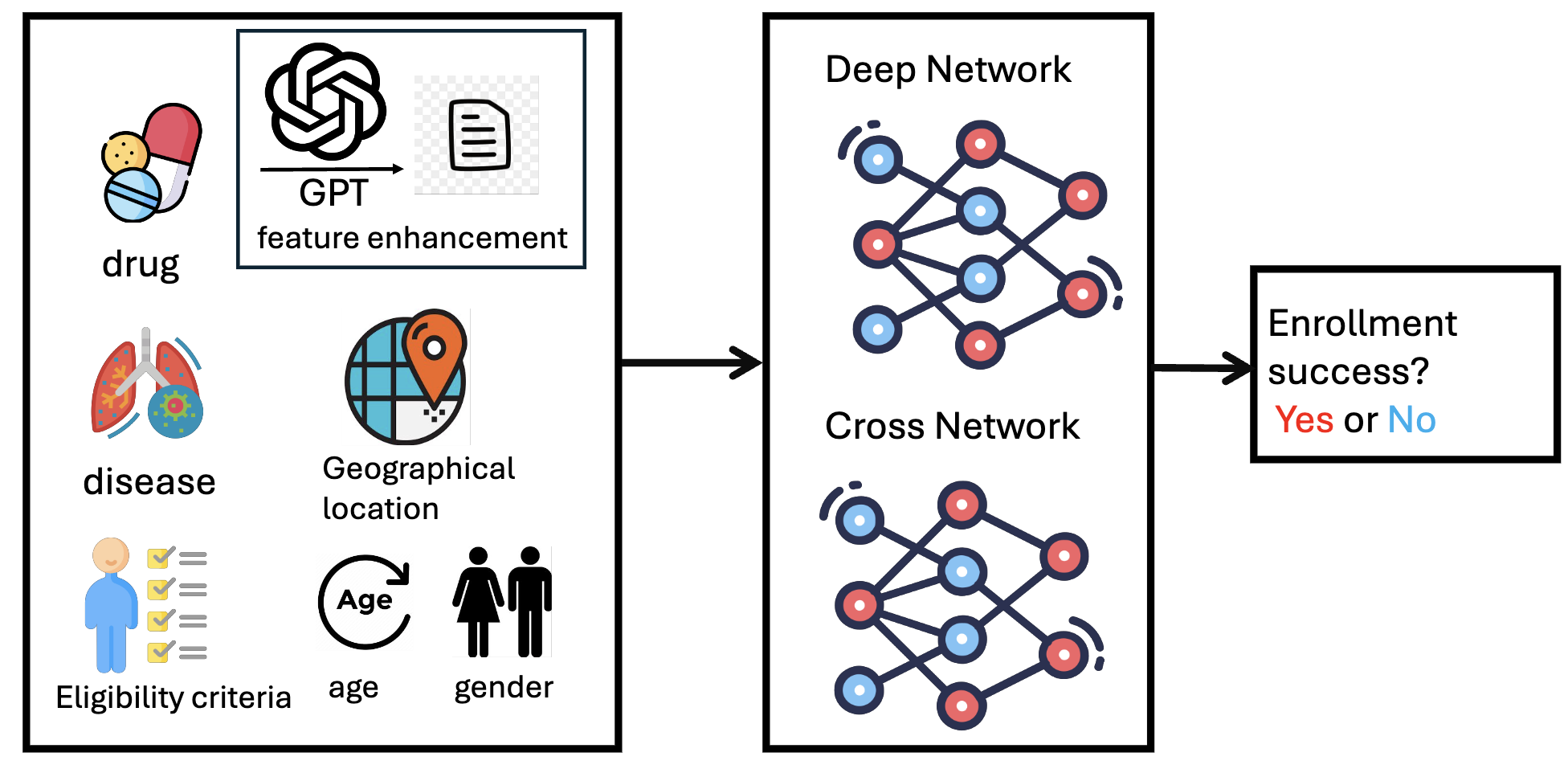}
\caption{Overview of TrialEnroll. Our model takes multimodal clinical trial features (e.g., drug, disease, eligibility criteria, geographical location of the trial, age, and target gender) as the input (detailed in Section~\ref{sec:feature_engineering}), augmented by large language model (Section~\ref{sec:llm_feature_enhancement}), leverages deep \& cross network (DCN) as neural architecture (Section~\ref{sec:dcn_feature_learning}), and predicts whether the trial enrollment will succeed. }
\label{fig:method}
\end{figure}

\noindent\textbf{Overview.}
In this section, we demonstrate the methodology. 
First, we discuss the broad impact of trial enrollment success prediction in Section~\ref{sec:broad_impact}. 
Then, we formulate the clinical trial enrollment success prediction problem in Section~\ref{sec:formulation}. Then, we discuss how to conduct feature engineering to produce informative features specifically for enrollment success prediction in Section~\ref{sec:feature_engineering}. 
After that, we describe how to leverage large language model (LLM) to augment the text feature in Section~\ref{sec:llm_feature_enhancement}. 
Then, we describe the customized Deep \& Cross Network in Section~\ref{sec:dcn_feature_learning}. 
For ease of understanding, the architecture of the whole model is illustrated in Figure~\ref{fig:method}.

\subsection{Broad Impact}
\label{sec:broad_impact}

Predicting clinical trial enrollment success is crucial for pharmaceutical companies~\cite{haddad2015barriers}. Accurate predictions enable effective resource allocation, guiding investments in time, money, and personnel. Enrollment delays or failures can significantly increase costs, but predictive models can identify potential issues early, allowing for adjustments that save substantial resources. These models also inform trial design, optimizing inclusion/exclusion criteria and trial locations to attract and retain participants.

Reliable enrollment predictions enhance stakeholder confidence~\cite{cruz2020impact}, including investors, partners, and regulatory bodies. Demonstrating a high likelihood of successful enrollment reinforces trust in the company's capabilities and commitment to delivering on its pipeline.

Accurate forecasting improves resource allocation and financial planning. By predicting enrollment outcomes, trial managers can optimize staffing, budget, and facilities, ensuring resources are available when needed and preventing inefficiencies and bottlenecks~\cite{Kerali2018}. This enhances budget accuracy and ensures efficient capital use, reducing waste and improving trial efficiency~\cite{Baskin2019}.

Predicting enrollment success is vital for proactive risk management. Early identification of potential recruitment challenges and delays allows for the development of contingency plans, minimizing impacts on the trial timeline and outcomes~\cite{Prasad2024}. Retrospective analyses of past trials and prospective data collection during ongoing trials further support this proactive approach~\cite{Knirsch2012}.

Effective communication with stakeholders is another critical benefit. Setting realistic expectations regarding trial timelines and outcomes fosters trust and collaboration with sponsors, regulatory bodies, and participants~\cite{Yu2024}. Reliable enrollment forecasts also help secure funding from sponsors and grant committees by providing accurate and trustworthy data.

For pharmaceutical companies, accurate forecasting of trial enrollment is essential for strategic planning, including market entry and product launch strategies. Enrollment predictions determine the timing of drug approval and market availability, crucial for competitive positioning and financial planning~\cite{dimasi2016innovation}. These forecasts support regulatory submission planning, facilitating smoother interactions with regulatory bodies and a more efficient approval process, ultimately speeding up market access for new treatments and providing a significant competitive advantage~\cite{ALSULTAN20201217}.

\subsection{Formulation of Clinical Trial Enrollment Success Prediction}
\label{sec:formulation}

A {\textit{clinical trial}} is a systematic effort to assess the safety and effectiveness of a specific set of {\it treatment set} designed to address a particular group of {\it target disease set}, This evaluation is conducted according to predefined {\it trial eligibility criteria} for a chosen group of patients.

\begin{definition}[\textbf{Drug Set}]
The drug set consists of a range of drug molecule candidates, denoted as $\mathcal{D} = \{ d_1, d_2, \ldots, d_N \}$, where $d_1, d_2, \ldots, d_N$ are the names of the $N$ drug molecules involved in this trial. This study focuses on trials that aimed at discovering new uses for these drug candidates while excluding trials that involve non-drug interventions such as surgery or medical devices.
\begin{equation}
\label{eqn:drug_set}
\mathcal{D} = \{d_1, d_2, \ldots, d_N\}.
\end{equation}
\end{definition}

\begin{definition}[\textbf{Target Disease Set}]
\label{def:disease}
For a trial addressing $K_{\delta}$ diseases, the Target Disease Set is represented by 
$\mathcal{T} = \{ t_1, t_2, \ldots, t_{K_{\delta}} \}$, with each $t_i$ being the disease name for the $i$-th disease.
\begin{equation}
\label{eqn:disease_advanced}
\mathcal{T} = \{ t_1, t_2, \ldots, t_{K_{\delta}} \}.
\end{equation}
\end{definition}

\begin{definition}[\textbf{Trial Eligibility Criteria}]
The trial eligibility criteria encompass both inclusion (+) and exclusion (-) criteria, which respectively outline the desired and undesirable attributes of potential participants. These criteria provide details on various key parameters such as age, gender, location, medical background, the status of the target disease, and the present health condition.
\begin{equation}
\label{eqn:protocol_advanced}
\mathcal{C} = [\bm{\psi}_1^+, \ldots, \bm{\psi}_Q^+, \bm{\psi}_1^-, \ldots, \bm{\psi}_R^-],\ \ \ \ \ \ \bm{\psi}_k^{+/-} \ \text{is a criterion}, 
\end{equation}
where $Q$ ($R$) is the number of inclusion (exclusion) criteria in the trial. The term $\bm{\psi}_k^+$ ($\bm{\psi}_k^-$) designates the $k$-th inclusion (exclusion) criterion within the eligibility criteria. Each criterion $\bm{\psi}$ is a sentence in unstructured natural language. 
\end{definition}

\begin{definition}[\textbf{Clinical Trial Categorical/Numerical Feature}]
Other clinical trial features also have considerable impacts on trial enrollment, including (1) demographic features, for example, the gender of the recruited patients (male or female or both), age of recruited patients (maximum and minimum age); (2) the trial phase (phase I, II, III, or IV); and (3) the geographical feature of the trial. 
These features are mostly categorical or numerical, denoted $\mathcal{W}$. 
\end{definition}

\begin{definition}[\textbf{Clinical Trial Enrollment Success}]
The trial enrollment success is the groundtruth of our model, denoted $y \in \{0,1\}$, a binary variable indicating the success of trial enrollment.     
\end{definition}

\noindent\textbf{Problem (Clinical Trial Enrollment Success Prediction).}
The estimation of \( y \), represented as \( \widehat{y} \), can be formulated through the machine learning model \( f_{\Theta} \), such that 
\begin{equation}
\widehat{y} = f_{\Theta}(\mathcal{D}, \mathcal{T}, \mathcal{C}, \mathcal{W}), 
\end{equation}
where \( \widehat{y}\) denotes the predicted enrollment state of a trial; $f_{\Theta}$ refers to the parameterized machine learning model, $\Theta$ denotes the learnable parameter. In this context, \( \mathcal{D} \), \( \mathcal{T} \), \( \mathcal{C} \) and $\mathcal{W}$ refer to the drug set, the target disease set, the trial eligibility criteria, and other categorical features, respectively. 
For ease of exposition, Table~\ref{table:trial_example} shows an example of a real clinical trial and all the related features. 

\begin{table*}[h!]
\centering
\caption{A real example of a clinical trial record.}
\begin{tabular}{c|p{1.65\columnwidth}}
\toprule
Feature & Descriptions \\ 
\midrule
NCTID & NCT00610792 (trial identifier) \\ 
drug &  bortezomib and pegylated liposomal doxorubicin \\ 
disease & Ovarian Cancer \\ 
phase & II \\ 
country & Italy, Switzerland \\
gender & Female \\
study type & interventional \\ 
title &  Phase 2 Study of Twice Weekly VELCADE and CAELYX in Patients With Ovarian Cancer Failing Platinum Containing Regimens \\ 
summary & This is a Phase 2, multicenter open-label, uncontrolled 2-step design.  Patients will be arranged in two groups based on the response to their last platinum containing therapy. \\
& The two groups are, 1) Platinum-Resistant Patients: patients with the progressive disease while on platinum-containing therapy or stable disease after at least 4 cycles; patients relapsing following an objective response while still receiving treatment; 
patients relapsing after an objective response within 6 months from the discontinuation of the last chemotherapy 
 and 2) Platinum-Sensitive Patients: patients who relapsed following an objective response \\ 
inclusion criteria & ECOG performance status grade 0 or 1
; Age $\geq$ 18 and $\leq$ 75 yrs; Life expectancy of at least 3 months; LVEF must be within normal limits; ...   
\\ 
exclusion Criteria & Chemotherapy, hormonal, radiation or immunotherapy or participation in any investigational drug study within 4 weeks of study entry; 
Pre-existing peripheral neuropathy $>$ Grade 1; 
Presence of cirrhosis or active or chronic hepatitis; ... 
Pregnancy or lactation or unwillingness to use adequate method of birth control; 
Active infection; 
Known history of allergy to mannitol, boron or liposomally formulated drugs. 
\\ 

start date & July 2006 \\ 
completed date & September 2009 \\ 
duration & 3.2 years \\ 
sponsor & Millennium Pharmaceuticals, Inc. \\ 
outcome & withdrawn \\ 
\bottomrule
\end{tabular}
\label{table:trial_example}
\end{table*}

\subsection{Feature Engineering}
\label{sec:feature_engineering}

In this section, we detail the handcrafted feature engineering process employed to transform raw clinical trial data into structured inputs suitable for machine learning models. The features are derived from various aspects of the clinical trial records, including drug information, disease characteristics, eligibility criteria, demographic details, and geographical information. For each categorical feature, we use one-hot encoding to represent it.

\subsubsection{Drug Embedding}
To represent the drug information, we utilize the pre-trained BioBERT model~\cite{lee2020biobert}, which is specifically trained on biological texts. A single clinical trial may involve several drugs. Each drug name is converted into an embedding vector using BioBERT. We then apply mean pooling to aggregate the token embeddings into a single fixed-size vector representing the drug. This approach captures the semantic information inherent in the drug names.

\subsubsection{Disease Embedding}
Similar to the drug embedding process, we use the pre-trained BioBERT model~\cite{lee2020biobert} to obtain embeddings for the disease names. A single clinical trial may involve several diseases. Each disease name is transformed into an embedding vector, and mean pooling is applied to generate a final disease embedding. This method ensures that the semantic nuances of the disease names are effectively captured.

\subsubsection{Eligibility Criteria Embedding}
The eligibility criteria, which include both inclusion and exclusion criteria, are processed at the sentence level. Each criterion sentence is embedded using the pre-trained BioBERT model~\cite{lee2020biobert}, specifically utilizing the [CLS] token to obtain the sentence embedding. This token is designed to capture the overall meaning of the sentence, making it suitable for representing the criteria.

\subsubsection{Demographic Features} 
\paragraph{Gender:} The gender of the participants is included as categorical features, with possible values being ``female'', ``male'', and ``all''. This categorical representation allows the model to distinguish between different gender requirements.

\paragraph{Age:} The minimum and maximum age requirements are included as numerical features. These values are critical as they influence the difficulty of enrolling participants within the specified age range.

\subsubsection{Trial Phase}
The phase of the clinical trial is treated as a categorical feature. 
There are four phases: Phase I (safety), Phase II (efficacy and side effects), Phase III (comparative effectiveness), and Phase IV (post-market surveillance). 
Each phase is encoded as a separate category, allowing the model to differentiate between the distinct stages of clinical development.

\subsubsection{Criteria Count}
We introduce a feature representing the count of inclusion and exclusion criteria, as well as the total number of criteria. The hypothesis is that a higher number of criteria may correlate with increased difficulty in enrolling participants.

\subsubsection{Geographical Features}
\paragraph{Country, State, City:} The geographical location of the trial is a categorical feature. Different regions may have varying distributions of disease prevalence and patient availability, making this a valuable feature.

\subsection{Large Language Model-based Feature Enhancement}
\label{sec:llm_feature_enhancement}

\begin{table*}[htbp]
\centering
\caption{Prompts that are used in LLM-based feature enhancement (Section~\ref{sec:llm_feature_enhancement}). }
\begin{tabular}{c|c|p{1.5\columnwidth}}
\toprule
\textbf{Category} & \textbf{Prompt Type} & \textbf{Description} \\ 
\midrule
\multirow{2}{*}{Drug} & System Prompt & You are a highly knowledgeable clinical pharmacologist. Given a string that contains the name of a drug, please:

- Provide the name of the drug. 

- Offer a comprehensive description of the drug, including:

-- Mechanism of action

-- Common uses

-- Notable side effects

-- Discuss the difficulty of recruiting patients for clinical trials involving this drug, including the reasons behind these challenges.

Noted instruction: Please respond with fewer than 100 words. \\ 
\cline{2-3}
 & User Prompt & The following string contains the name of a drug: <string>\{drug\}</string> \\ 
\midrule
\multirow{2}{*}{Disease} & System Prompt &  You are a highly knowledgeable clinical epidemiologist. Given a string that contains the name of a disease, please:

- Provide the name of the disease.

- Offer a comprehensive description of the disease, including:

-- Pathogenesis (mechanism of disease development) 

-- Common symptoms

-- Typical treatment options

-- Discuss the difficulty of recruiting patients for clinical trials involving this disease, including the reasons behind these challenges.

Noted instruction: Please respond with fewer than 100 words.\\ 
\cline{2-3}
 & User Prompt & The following string contains the name of a disease: <string>\{disease\}</string> \\ 
\bottomrule
\end{tabular}
\label{table:prompt}
\end{table*}

Large Language Models (LLMs) are advanced artificial intelligence systems~\cite{zhao2023survey} designed to process, understand, generate, and manipulate natural language text. Leveraging the vast amounts of data they are trained on, LLMs can significantly enhance feature representation in various applications, including clinical trial data analysis.

In our approach, we utilize LLMs to enrich the representations of drug and disease information beyond the capabilities of traditional embeddings. The process involves two main steps: generating detailed contextual information and embedding this information using BioBERT~\cite{lee2020biobert}.

\subsubsection{Enhanced Drug and Disease Representation}
To enhance the representation of drugs and diseases, we first generate comprehensive introductory paragraphs for each entity using a large language model (LLM)~\cite{jiang2023mistral}\footnote{Model version: mistralai/Mistral-7B-Instruct-v0.3.}. This approach leverages the LLM's ability to synthesize relevant knowledge from its extensive training data, providing a richer context for each drug and disease.

\paragraph{Step 1: Context Generation}
For each drug and disease, we prompt the LLM with a specific query to generate an introductory paragraph. The prompt is designed to elicit detailed information that captures the essential characteristics and context of the drug or disease. Table~\ref{table:prompt} displays the prompt.

\paragraph{Step 2: Embedding Generation}
Once the introductory paragraphs are generated, we use the pre-trained BioBERT model to obtain embeddings for these paragraphs. Specifically, we utilize the [CLS] token to capture the overall meaning of the text. This token is designed to aggregate the contextual information of the entire paragraph into a single embedding vector.

\paragraph{Step 3: Integration into the Model}
The resulting embeddings from the introductory paragraphs are then integrated into our model. These enhanced embeddings provide a richer and more nuanced representation of the drugs and diseases, potentially improving the model's performance in predicting the success of clinical trial enrollment.

By integrating LLM-generated contextual information and leveraging their analytical capabilities, we significantly enhance the feature representation and analysis processes, leading to more accurate and insightful predictions for predicting the success of clinical trial enrollment.

\begin{figure}[htbp]
\centering
\includegraphics[width=0.47\textwidth]{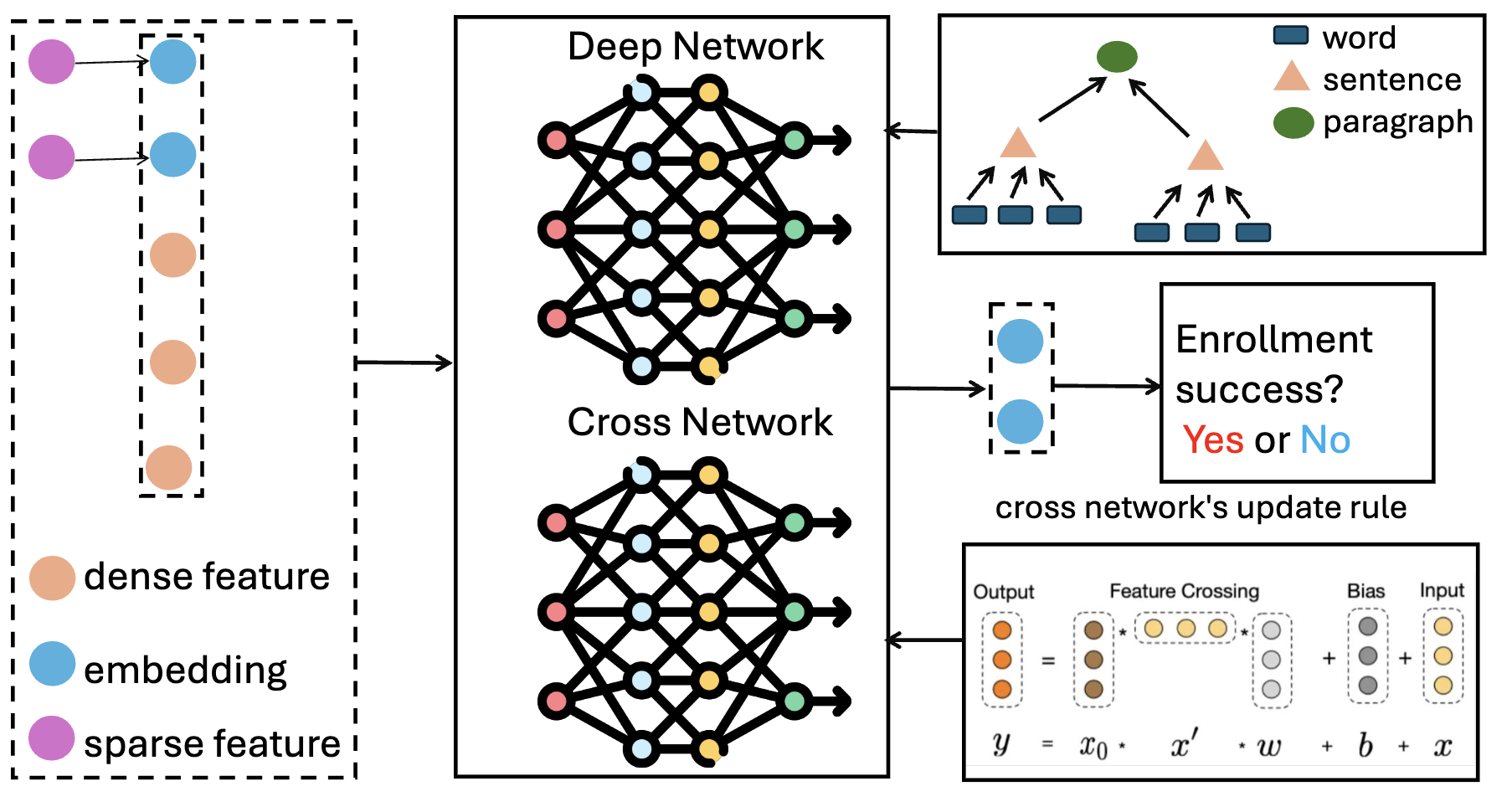}
\caption{The Deep \& Cross Network. }
\label{fig:model_arc}
\end{figure}

\subsection{Deep \& Cross Network}
\label{sec:dcn_feature_learning}

After obtaining the features, we customize  Deep \& Cross Network (DCN)~\cite{wang2017deep} to effectively learn and integrate semantic information from both handcrafted features and embeddings derived from natural language processing (NLP) models. The model architecture is shown in Figure~\ref{fig:model_arc}.
The DCN consists of two essential components: a deep network and a cross-network. DCN explicitly models feature interactions at different levels. The cross-network component captures feature crosses efficiently, which can be more effective than the implicit feature interactions learned by MLPs. 

\subsubsection{Deep Network: Hierarchical Attention Network}
The deep network component of the DCN is implemented as a Hierarchical Attention Network (HAN), inspired by the architecture used in~\cite{yue2024trialdura}. The HAN is designed to capture hierarchical structures in the eligibility criteria data, such as the relationships between words and sentences. This is particularly useful for processing the rich textual information embedded in clinical eligibility trial data. 

The hierarchical attention mechanism operates at two levels:
\begin{itemize}[leftmargin=*]
    \item \textbf{Word-level attention:} This layer captures the importance of each word within a sentence, allowing the model to focus on the most relevant words.
    \item \textbf{Sentence-level attention:} This layer captures the importance of each sentence within the inclusion (or exclusion) criteria, enabling the model to focus on the most relevant criteria sentences. 
    Sentence-level attention is of particular importance in clinical trials. 
    For example, considering two inclusion criteria: one is to recruit patients with Amyotrophic Lateral Sclerosis (ALS), and the other is to recruit patients aged 18 or older. 
    Clearly, due to the rarity of ALS, the criterion related to ALS has a more significant impact.
\end{itemize}

Formally, let $\mathbf{h}_{it}$ be the hidden state of the $t$-th word in the $i$-th sentence. The word-level attention mechanism computes a context vector $\mathbf{u}_i$ for each sentence as follows:
\begin{equation}
    \mathbf{u}_i = \sum_{t} \alpha_{it} \mathbf{h}_{it},
\end{equation}
where $\alpha_{it}$ are the attention weights that determine the importance of each word in the sentence.

Similarly, the sentence-level attention mechanism computes a document-level context vector $\mathbf{u}$ as:
\begin{equation}
    \mathbf{u} = \sum_{i} \beta_{i} \mathbf{u}_i,
\end{equation}
where $\beta_{i}$ are the attention weights that determine the importance of each sentence in the document.

\subsubsection{Cross Network}
The cross network component is designed to explicitly learn feature interactions, particularly those involving handcrafted features. Unlike MLPs, which may struggle to capture higher-order interactions, the cross network efficiently models these interactions through a series of cross layers.

Given an input feature vector $\mathbf{x} \in \mathbb{R}^d$, the cross network computes the $l$-th cross layer as:
\begin{equation}
    \mathbf{x}^{(l+1)} = \mathbf{x}^{(0)} \mathbf{x}^{(l)^\top} \mathbf{w}^{(l)} + \mathbf{b}^{(l)} + \mathbf{x}^{(l)},
\end{equation}
where $\mathbf{x}^{(0)}$ is the original input vector, $\mathbf{w}^{(l)}$ and $\mathbf{b}^{(l)}$ are the weights and biases of the $l$-th cross layer, respectively. This formulation allows the model to learn explicit feature interactions iteratively.

\subsubsection{Integration and Training}
The outputs of the deep network (HAN) and the cross network are concatenated and fed into a final fully connected layer. This combined representation leverages both the hierarchical semantic information captured by the deep network and the explicit feature interactions learned by the cross network.

The final prediction $\widehat{y}$ is computed as:
\begin{equation}
    \widehat{y} = \sigma\big(\mathbf{w}_p^\top [\mathbf{u}; \mathbf{x}^{(L)}] + b_p\big),
\end{equation}
where $\sigma(\cdot)$ is the activation function (e.g., sigmoid for binary classification), $\mathbf{w}_p$ and $b_p$ are the weights and bias of the final fully connected layer, and $L$ is the number of cross layers.
We utilize binary cross-entropy loss as the loss criterion. 
The objective function is 
\begin{equation}
\sum_{i=1}^N - y_i \log \widehat{y}_i - (1-y_i)\log (1-\widehat{y}_i),     
\end{equation}
where $N$ is the number of data points in training set, $y_i$ and $\widehat{y}_i$ are groundtruth and prediction of the $i$-th data point, respectively. 
AdamW~\cite{loshchilov2017decoupled} is used as the numerical optimizer to minimize the objective function. 

By integrating the hierarchical attention network and the cross network, the customized DCN effectively captures both deep semantic information and explicit feature interactions, leading to improved performance in predicting the enrollment success of a clinical trial.

\section{Experiment} 
\label{sec:experiment}

In this section, we present the empirical studies. 
We first describe the data curation process in Section~\ref{sec:data}. 
Then, Section~\ref{sec:setup} briefly describes the experimental setup. 
After that, we present the experimental results and ablation studies in Section~\ref{sec:experimental_results} and \ref{sec:ablation}.

\subsection{Data Curation}
\label{sec:data}
For this study, we utilized a dataset from ClinicalTrials.gov~\cite{zarin2011clinicaltrials}, a comprehensive global registry of clinical trials. Our objective was to analyze the factors influencing the enrollment success of these trials. We specifically focused on trials with complete datasets to ensure the reliability of our analysis.

Each clinical trial in the dataset was represented as an XML file. From these files, we extracted a variety of pertinent information, including the National Clinical Trial (NCT) ID, geographical location, gender, age, trial phase, disease name, drug name, and eligibility criteria (both inclusion and exclusion criteria). Following data extraction, we conducted feature engineering and incorporated features enhanced by large language models (LLMs).

To prevent data leakage, we divided the dataset into training and testing subsets based on a temporal cutoff of January 1, 2015. 
Trials that concluded before this date were included in the training set, while those that commenced after this date were allocated to the testing set. 
This temporal split ensured that the training data did not contain information from the future relative to the testing data. 
Our final dataset comprised 31,094 records, 22,579 of which were allocated to the training set and 8,515 to the testing set. 
The distribution of trial completion counts by year range is detailed in Table~\ref{tab:completion_year_counts}.

\begin{table*}[htbp]
\small 
\centering
\caption{Distribution of trial completion counts by year range.}
\begin{tabular}{ccccccc}
\toprule
Year range & 
Before 2000 &
2000-2004     & 
2005-2009     & 
2010-2014     & 
2015-2019     & 
After 2020    \\ 
\midrule 
\# trials & 52 & 1,055 & 10,133 & 11,643 & 7,969 & 242 \\ 
\bottomrule
\end{tabular}
\label{tab:completion_year_counts}
\end{table*}

Given the imbalanced nature of the dataset, where the distribution is heavily skewed towards class 0 (non-enrollment) as shown in Table~\ref{tab:class_distribution}, it is crucial to comprehensively assess the performance metrics to understand the strengths and weaknesses of each model. To address this imbalance, we employed oversampling techniques to increase the number of samples in the minority class (class 1) to match the number of samples in the majority class (class 0).

\begin{table}[htbp]
\small 
\centering
\caption{Distribution of clinical trial enrollment success by phase.}
\begin{tabular}{ccccc}
\toprule
Phase & I & II & III & IV \\ \midrule 
\# negative & 5,564 & 11,297 & 7,227 & 3,698 \\
\# positive & 426 & 1,284 & 422 & 496 \\ 
\bottomrule
\end{tabular}
\label{tab:class_distribution}
\end{table}

\begin{figure}[htbp]
\centering
\includegraphics[width=0.47\textwidth]{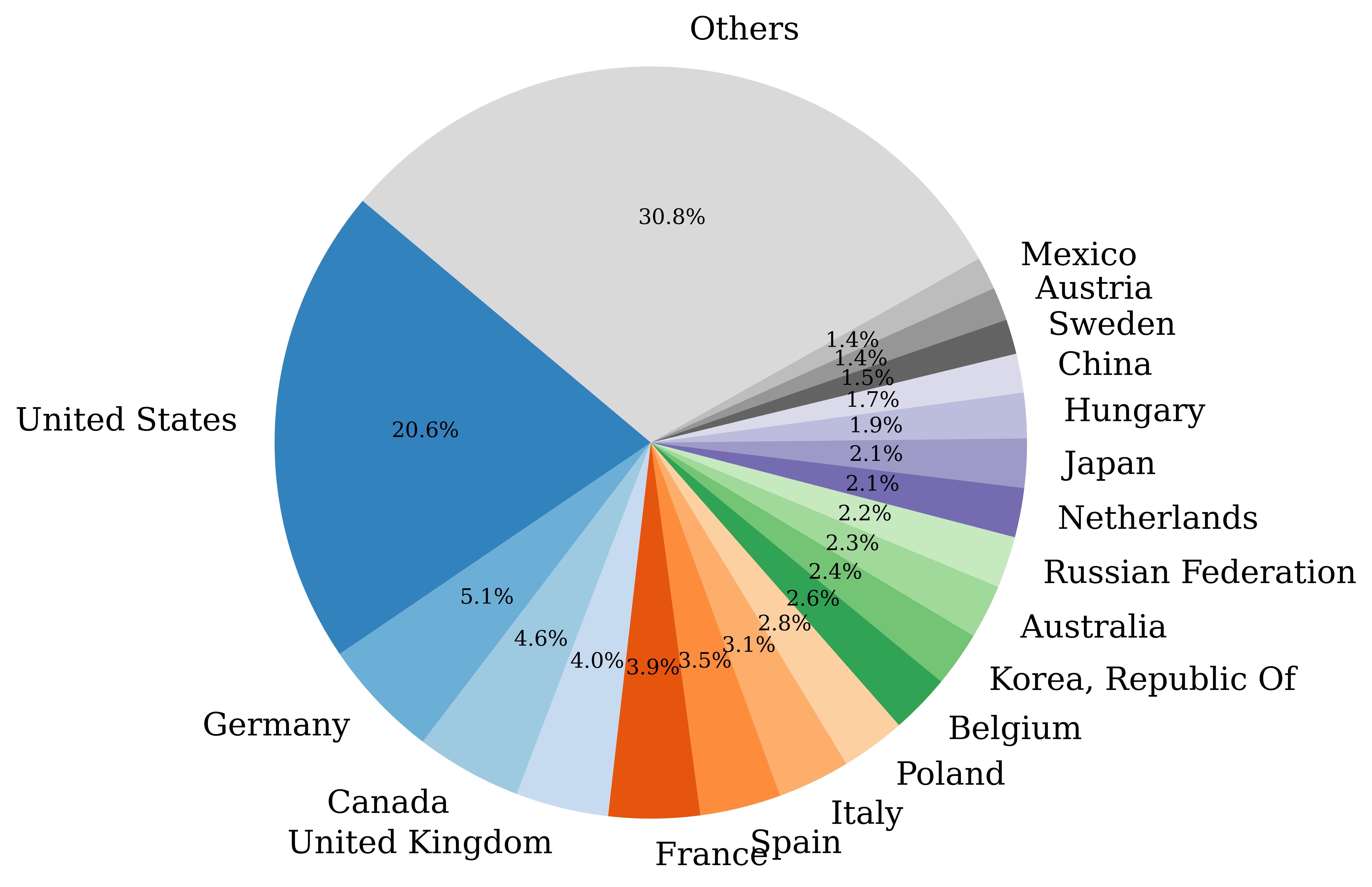}
\caption{Distribution of geographical locations of clinical trial records (country-level). }
\label{fig:country_distribution}
\end{figure}

The pie chart in Figure~\ref{fig:country_distribution} illustrates the distribution of clinical trial records by country. The United States has the largest share of records, comprising 20.6\% of the total. This is followed by Germany (5.1\%), Canada (4.6\%), the United Kingdom (4.0\%), and France (3.9\%). Other notable countries include Spain, Italy, Belgium, Poland, and several others, each contributing between 1.4\% and 3.5\%. The remaining countries are grouped into the ``Others'' category, which makes up 30.8\% of the records. This distribution highlights the global nature of clinical trials but also indicates a concentration in certain countries.

Table~\ref{tab:age_feature_statistics} presents a summary of the statistics for the age-related features used in our model. The features include Min-age, Max-age, and Age-span. Min-age represents the minimum age recorded, Max-age denotes the maximum age recorded, and Age-span is the range between the minimum and maximum ages. Note that the minimum value recorded is -1, which indicates missing or undefined values.

\begin{table}[htbp]
\small 
\centering
\caption{Summary statistics of age-related features. }
\begin{tabular}{lcccc}
\toprule
Statistic & Min-age & Max-age & Age-span \\
\midrule
Mean & 19.26 & 29.42 & 20.24 \\
Min & -1.00 & -1.00 & -66.00 \\
25\% & 18.00 & -1.00 & -1.00 \\
50\% & 18.00 & -1.00 & -1.00 \\
75\% & 18.00 & 65.00 & 47.00 \\
Max & 83.00 & 365.00 & 132.00 \\
\bottomrule
\end{tabular}
\label{tab:age_feature_statistics}
\end{table}

Table~\ref{tab:sentence_feature_statistics} provides a summary of the statistics for the sentence-related features used in our clinical trial enrollment prediction model. The features include Inclusion Criteria Count, Exclusion Criteria Count, and Total Criteria Count, which is the sum of the inclusion and exclusion criteria.

\begin{table}[htbp]
\small 
\centering
\caption{Summary statistics of Criteria Count. }
\begin{tabular}{lccc}
\toprule
Statistic & Inclusion & Exclusion & Total \\
\midrule
Mean & 13.16 & 18.20 & 31.36 \\
Min & 0.00 & 0.00 & 0.00 \\
25\% & 4.00 & 6.00 & 12.00 \\
50\% & 9.00 & 13.00 & 23.00 \\
75\% & 17.00 & 25.00 & 44.00 \\
Max & 163.00 & 174.00 & 215.00 \\
\bottomrule
\end{tabular}
\label{tab:sentence_feature_statistics}
\end{table}

Table~\ref{tab:gender_distribution} shows the distribution of gender within the dataset. ``All genders'' means both female and male participants can be recruited.

\begin{table}[htbp]
\centering
\caption{Distribution of gender in the dataset. }
\begin{tabular}{lcc}
\toprule
Gender & Count & Proportion (\%) \\
\midrule
All genders & 27,316 & 87.85 \\
Female & 2,450 & 7.87 \\
Male & 1,328 & 4.27 \\
\bottomrule
\end{tabular}
\label{tab:gender_distribution}
\end{table}

\begin{table*}[ht]
\centering
\caption{Summary of model performance for clinical trial enrollment success prediction. Results are averaged over three independent runs; corresponding standard deviations are also shown. }
\begin{tabular}{ccccccc}
\toprule
Model & PR-AUC & ROC-AUC & F1 score & Precision & Recall & Accuracy \\
\midrule
LR & 0.6499 $\pm$ 0.0027 & 0.6814 $\pm$ 0.0035 & 0.4463 $\pm$ 0.0058 & 0.6816 $\pm$ 0.0023 & 0.3318 $\pm$ 0.0060 & 0.5884 $\pm$ 0.0023 \\
GBDT & 0.6660 $\pm$ 0.0102 & 0.6317 $\pm$ 0.0010 & \textbf{0.6191 $\pm$ 0.0135} & 0.5960 $\pm$ 0.0025 & \textbf{0.6445 $\pm$ 0.0259} & 0.6039 $\pm$ 0.0064 \\
AdaBoost & 0.6525 $\pm$ 0.0022 & 0.6816 $\pm$ 0.0029 & 0.5631 $\pm$ 0.0062 & 0.6552 $\pm$ 0.0021 & 0.4937 $\pm$ 0.0085 & \textbf{0.6169 $\pm$ 0.0031} \\
RF & 0.6725 $\pm$ 0.0042 & 0.6796 $\pm$ 0.0039 & 0.3739 $\pm$ 0.0167 & \textbf{0.7591 $\pm$ 0.0058} & 0.2482 $\pm$ 0.0142 & 0.5848 $\pm$ 0.0060 \\
MLP & 0.6773 $\pm$ 0.0060 & 0.7146 $\pm$ 0.0028 & 0.4938 $\pm$ 0.0907 & 0.6975 $\pm$ 0.0259 & 0.4002 $\pm$ 0.1316 & 0.6094 $\pm$ 0.0239 \\
\textbf{TrialEnroll} & \textbf{0.7002 $\pm$ 0.0013} & \textbf{0.7352 $\pm$ 0.0021} & 0.4507 $\pm$ 0.0529 & 0.7412 $\pm$ 0.0102 & 0.3275 $\pm$ 0.0567 & 0.6060 $\pm$ 0.0160 \\
\bottomrule
\end{tabular}
\label{tab:model_performance_enrollment}
\end{table*}

\subsection{Experimental Setup}
\label{sec:setup}

In this section, we briefly describe the experimental setup, including evaluation metrics, baseline methods and implementation details. A detailed description is available in the Appendix. 

\noindent\textbf{Evaluation metrics.}
Clinical trial enrollment success prediction is formulated as a binary classification in this paper. 
We use six different evaluation metrics to measure performance comprehensively, including Precision-Recall Area Under Curve (PR-AUC), Area Under the Receiver Operating Characteristic Curve (ROC-AUC), F1 score, precision, recall, and accuracy. The scores for all six metrics range from 0 to 1; a higher value represents better performance.

\noindent\textbf{Baseline methods.}
We selected multiple widely recognized models as baselines, including Logistic Regression (LR), Gradient Boosting Decision Tree~\cite{ke2017lightgbm} (GBDT), Adaptive Boosting~\cite{ratsch2001soft} (AdaBoost), Random Forest~\cite{breiman2001random} (RF), and Multi-Layer Perceptron~\cite{popescu2009multilayer} (MLP). The data fed into each model is the same to ensure a fair comparison.

\noindent\textbf{Implementation details} are elaborated in Section~\ref{sec:implement} in Appendix.

\subsection{Experimental Results}
\label{sec:experimental_results}

As illustrated in Table~\ref{tab:model_performance_enrollment}, the performance of various predictive models for clinical trial enrollment success prediction is evaluated using metrics such as PR-AUC, ROC-AUC, F1 Score, Accuracy, Precision, and Recall.

Among all the models evaluated, TrialEnroll demonstrates the best performance on the PR-AUC, which is the most critical metric in clinical trials. This model achieves a PR-AUC of 0.7002 and an ROC-AUC of 0.7352, both of which surpass those of the baseline models, indicating its robustness in accurately predicting trial enrollment success.

However, it is noteworthy that GBDT achieves the highest scores in F1 Score and Recall, while Random Forest excels in the Precision metric, and AdaBoost stands out in terms of Accuracy. GBDT, AdaBoost, and Random Forest are particularly well-suited for tabular data, and their superior performance further underscores the effectiveness of our handcrafted feature engineering and the incorporation of LLM-enhanced features. TrialEnroll does not outperform all models across every metric. Therefore, in practical applications, users should consider model complexity and specific task metrics to select the most appropriate model.

Importantly, this paper is the first to identify the problem of predicting trial enrollment success, design a novel feature engineering approach incorporating LLM features, and provide a benchmark for several well-established machine learning models. While the novel model TrialEnroll excels in PR-AUC and ROC-AUC, it does not achieve the best performance across all metrics. Future work will focus on designing an improved model to address this limitation.

To further demonstrate the importance of handcrafted features, we used permutation importance with a simple logistic regression model. This method involves shuffling the values of each feature and measuring the drop in model performance (using PR-AUC) compared to the baseline. The difference in PR-AUC before and after shuffling quantifies the feature's importance. We repeated this process three times to obtain the mean importance scores and their standard deviations.

\begin{figure}[htbp]
\centering
\includegraphics[width=0.48\textwidth]{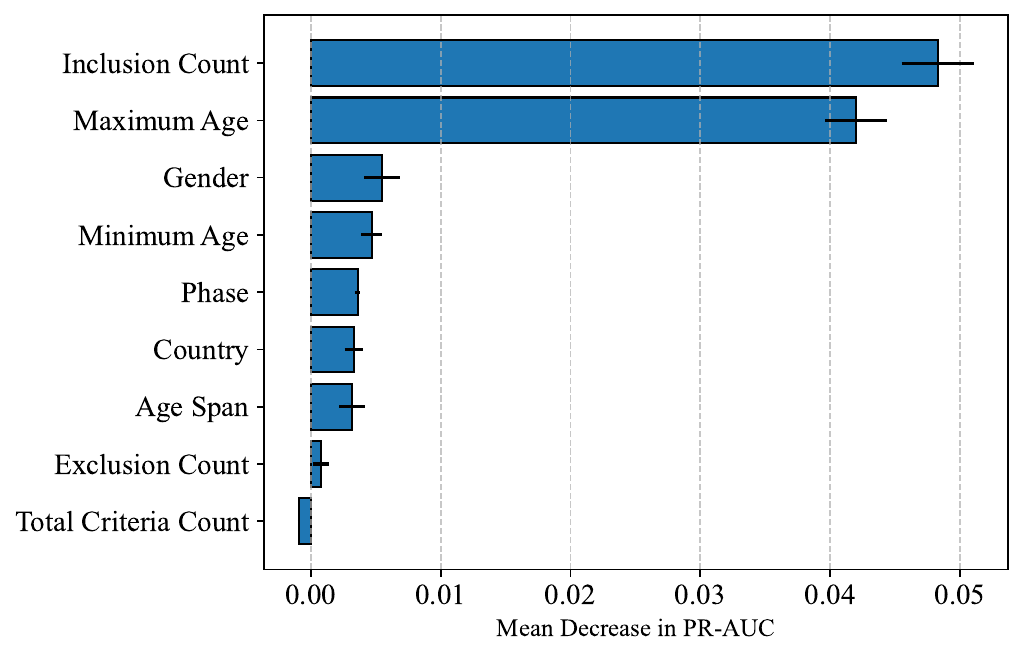}
\caption{Ablation study on feature. Permutation importance of features using PR-AUC. }
\label{fig:feature_importance}
\end{figure}

The results are shown in Figure~\ref{fig:feature_importance}. From the results, we can see that ``inclusion criteria count'' and ``max age'' are the most impactful features. This finding makes intuitive sense: the more inclusion criteria there are, the harder it is to enroll participants. Additionally, a higher maximum age may facilitate easier enrollment.


\begin{table*}[ht]
\centering
\caption{Comparison of enhanced features using MLP model.}
\label{tab:enhanced_features_comparison}
\resizebox{1.98\columnwidth}{!}{
\begin{tabular}{lcccccc}
\toprule
Method & PR-AUC  & ROC-AUC  & Accuracy  & F1  & Precision  & Recall  \\ 
\midrule
Origin & $0.6497 \pm 0.0173$ & $0.6864 \pm 0.0138$ & $0.6070 \pm 0.0365$ & \textbf{$0.5240 \pm 0.1428$} & $0.6549 \pm 0.0325$ & \textbf{$0.4792 \pm 0.1945$} \\
Origin + LLM & $0.6544 \pm 0.0116$ & $0.6887 \pm 0.0076$ & \textbf{$0.6108 \pm 0.0244$} & $0.5231 \pm 0.0960$ & $0.6715 \pm 0.0256$ & $0.4514 \pm 0.1464$ \\
Origin + LLM + Handcrafted & \textbf{0.6773 $\pm$ 0.0060} & \textbf{0.7146 $\pm$ 0.0028} & $0.6094 \pm 0.0239$ & $0.4938 \pm 0.0907$ & \bf{$0.6975 \pm 0.0259$} & $0.4002 \pm 0.1316$ \\
\bottomrule
\end{tabular}
}
\end{table*}

\subsection{Ablation Study}
\label{sec:ablation}

Table~\ref{tab:enhanced_features_comparison} presents an ablation study comparing the performance of different feature sets. The feature sets evaluated are: 
\begin{itemize}[leftmargin=*]
\item \textbf{Origin}: Includes embeddings for drug names, disease names, and criteria (both inclusion and exclusion).
\item \textbf{Origin + LLM}: Combines the Origin embeddings with additional features enhanced by a Large Language Model (LLM).
\item \textbf{Origin + LLM + Handcrafted}: Integrates the Origin and LLM-enhanced features with additional handcrafted features.
\end{itemize}

The results demonstrate that both LLM-enhanced features and handcrafted features contribute to significant improvements. The Origin feature set achieves a PR AUC of 0.6497. Adding LLM-enhanced features increases the PR AUC to 0.6544, and incorporating handcrafted features further improves it to 0.6773.

\section{Conclusion}
\label{sec:conclusion}

In this paper, we have addressed the problem of predicting clinical trial enrollment success by establishing a benchmark to evaluate well-known machine learning models and designing a customized Deep \& Cross Network (DCN) model named \textit{TrialEnroll}. Our approach also involved effective handcrafted feature engineering techniques.

\paragraph{Technical Contributions}
Our work presents several key technical contributions:
\begin{itemize}[leftmargin=*]
    \item \textbf{Benchmark Establishment:} We created a robust benchmark for evaluating machine learning models on clinical trial enrollment success prediction.
    \item \textbf{Customized DCN Model:} We designed \textit{TrialEnroll}, a DCN model that integrates a Hierarchical Attention Network (HAN) and a cross network to combine deep semantic information with handcrafted feature interactions.
    \item \textbf{Feature Engineering:} We developed a detailed handcrafted feature engineering process, including the use of pre-trained BioBERT embeddings for drug and disease information.
    \item \textbf{LLM-based Feature Enhancement:} We leveraged large language models (LLMs) to enrich drug and disease representations, improving model performance.
\end{itemize}

\paragraph{Clinical Implications}
The findings of this study have significant clinical implications:
\begin{itemize}[leftmargin=*]
    \item \textbf{Improved Enrollment Predictions:} Our model can help researchers and sponsors identify potential challenges early in the trial design process.
    \item \textbf{Resource Optimization:} Enhanced prediction capabilities enable more efficient allocation of resources, leading to more successful and cost-effective clinical trials.
    \item \textbf{Patient Recruitment:} Our model can assist in identifying trials that are more likely to succeed in enrolling participants.
\end{itemize}

\paragraph{Limitations and Future Work}
While our study presents promising results, there are several limitations and areas for future work:
\begin{itemize}[leftmargin=*]
    \item \textbf{Data Limitations:} Future work could involve expanding the dataset to include a more diverse range of trials.
    \item \textbf{Feature Expansion:} Incorporating additional features, such as patient-level data, could enhance predictive capabilities.
\end{itemize}

In summary, our study provides a comprehensive approach to predicting clinical trial enrollment success, offering valuable insights and tools for researchers and sponsors. The proposed \textit{TrialEnroll} model, along with our benchmark and feature engineering techniques, represents a significant step forward in the field of clinical trial optimization.

\small 
\bibliographystyle{ACM-Reference-Format}
\bibliography{reference}

\clearpage
\appendix 

\section{Additional Experimental Details}

In this section, we present the additional experimental details to enhance the reproducibility. Concretely, Section~\ref{sec:metric} details the evaluation metrics to assess model performance. 
Section~\ref{sec:baseline} elaborates on baseline methods. 
Section~\ref{sec:implement} describes the implementation details. 

\subsection{Evaluation Metrics}
\label{sec:metric}
Clinical trial enrollment success prediction is formulated as a binary classification in this paper. 
In binary classification, there are four kinds of test data points based on their ground truth and the model's prediction, 
(1) positive sample and is correctly predicted as positive, also known as \textit{True Positive} (TP);
(2) negative samples and is wrongly predicted as positive samples, also known as \textit{False Positive (FP)};
(3) negative samples and is correctly predicted as negative samples, also known as \textit{True Negative (TN)};
(4) positive samples and is wrongly predicted as negative samples, also known as \textit{False Negative (FN)}.

We use different evaluation metrics as follows. 
(1) {PR-AUC} (Precision-Recall Area Under Curve). The area under the Precision-Recall curve summarizes the trade-off between the true positive rate and the positive predictive value for a predictive model using different probability thresholds. 
(2) {ROC-AUC}. Area Under the Receiver Operating Characteristic Curve (ROC-AUC) summarizes the trade-off between the true positive rate and the false positive rate for a predictive model using different probability thresholds. ROC-AUC is also known as the Area Under the Receiver Operating Characteristic curve (AUROC) in some literature. 
(3) {F1}. The F1 score is the harmonic mean of the precision and recall, defined as 
$\text{F1} = \frac{2}{\frac{1}{\text{precision }} + \frac{1}{\text{recall}}}$. 
(4) {Precision}. The precision is the performance of a classifier on the samples that are predicted as positive. It is formally defined as $\text{precision} = \frac{TP}{TP+FP}$. 
(5) {Recall}. The recall score measures the performance of the classifier to find all the positive samples.
It is formally defined as $\text{recall} = \frac{TP}{TP+FN}$. 
(6) {Accuracy}. Accuracy is the fraction of correctly predicted/classified samples. 
It is formally defined as $\text{accuracy} = \frac{TP+TN}{TP+TN+FP+FN}$. 

For all these metrics, the numerical values range from 0 to 1, a higher value represents better performance. We report multiple metrics to measure the performance comprehensively. 

\subsection{Baseline Methods}
\label{sec:baseline}
We are the first to identify the Enrollment Success prediction problem. To address this problem, we propose a benchmark to evaluate performance using several widely recognized models alongside our customized Deep Cross Network~\cite{wang2017deep} (DCN), which we refer to as TrialEnroll.

To establish this benchmark, we selected multiple widely recognized models as baselines: Logistic Regression~\cite{lavalley2008logistic} (LR), Gradient Boosting Decision Tree~\cite{ke2017lightgbm} (GBDT), Adaptive Boosting~\cite{ratsch2001soft} (AdaBoost), Random Forest~\cite{breiman2001random} (RF), and Multi-Layer Perceptron~\cite{popescu2009multilayer} (MLP). The data fed into each model is the same to ensure a fair comparison. These models have been successfully applied to clinical trial outcome prediction in previous studies~\citep{fu2022hint,fu2023automated,lu2024uncertainty}.

\begin{itemize}[leftmargin=*]
    \item \textbf{Logistic Regression (LR)}: Logistic Regression is a simple and widely used statistical method for modeling the relationship between a dependent variable and one or more independent variables. It assumes a linear relationship between the input features and the output, making it easy to interpret but potentially limited in capturing complex patterns.
    
    \item \textbf{Gradient Boosting Decision Tree (GBDT)}: GBDT is an ensemble learning technique that builds a series of decision trees, where each tree corrects the errors of the previous ones. It combines the predictions of multiple weak learners to produce a strong learner, making it highly effective for both regression and classification tasks.
    
    \item \textbf{Adaptive Boosting (AdaBoost)}: AdaBoost is another ensemble learning method that combines multiple weak classifiers to form a strong classifier. It works by iteratively adjusting the weights of incorrectly classified instances, focusing more on difficult cases in subsequent iterations. This adaptive approach helps improve the overall model performance.
    
    \item \textbf{Random Forest (RF)}: Random Forest is an ensemble learning method that constructs multiple decision trees during training and outputs the mode of the classes (classification) or mean prediction (regression) of the individual trees. It reduces overfitting by averaging multiple trees, providing robust and accurate predictions.
    
    \item \textbf{Multi-Layer Perceptron (MLP)}: MLP is a type of artificial neural network that consists of multiple layers of nodes, including an input layer, one or more hidden layers, and an output layer. Each node (neuron) in a layer is connected to every node in the subsequent layer, allowing the network to learn complex, non-linear relationships in the data.
\end{itemize}

We aim to demonstrate TrialEnroll's effectiveness by comparing It with these well-established models. This benchmark will also provide insights into the relationship between model complexity and performance.

\subsection{Implementation Details}
\label{sec:implement}

To ensure reproducibility, we provide a detailed description of our experimental framework and training process. The code and step-by-step instructions can be found at \url{https://anonymous.4open.science/r/TrialEnroll-7E12}.

\paragraph{Hardware and Software Configuration}
\begin{itemize}
    \item \textbf{CPU:} Intel(R) Xeon(R) Gold 6248
    \item \textbf{RAM:} 128GB
    \item \textbf{GPU:} NVIDIA RTX A5000
    \item \textbf{Operating System:} Ubuntu 20.04
    \item \textbf{Python Version:} 3.10
    \item \textbf{PyTorch Version:} 2.3
\end{itemize}


\paragraph{Training Parameters}

During the training process, we set the batch size to 256 to balance memory usage and computational efficiency. The training was conducted over 100 epochs, with an early stopping mechanism to prevent overfitting. The early stopping patience was set to 5 epochs, meaning training would halt if no improvement in the validation loss was observed for 5 consecutive epochs.

\paragraph{Optimization}

We employed the AdamW optimizer~\cite{loshchilov2017decoupled} to minimize the objective function. The initial learning rate was set to 0.001, which we found to be effective for our model and dataset. The learning rate was not dynamically adjusted during training.

The entire training process took approximately 4.3 hours to complete. This duration may vary depending on the specific hardware configuration and the complexity of the model.

\end{document}